\documentclass{article}
\usepackage{spconf,amsmath,graphicx}
\usepackage{adjustbox}

\usepackage{enumitem}
\setlist{nosep, leftmargin=14pt}

\usepackage{mwe} % to get dummy images

\title{Benchmarking Computational Pathology
Foundation Models for Semantic Segmentation}
\name{Lavish Ramchandani, Aashay Tinaikar, Dev Kumar Das, Rohit Garg and Tijo Thomas}
\address{Aira Matrix Private Limited, Mumbai, India}
\begin{document}
%\ninept
%
\maketitle
\begin{abstract}

In recent years, foundation models such as CLIP, DINO, and CONCH have demonstrated remarkable domain generalization and unsupervised feature extraction capabilities across diverse imaging tasks. However, systematic and independent evaluations of these models for pixel-level semantic segmentation in histopathology remain scarce. In this study, we propose a robust benchmarking approach to asses 10 foundational models on four histopathological datasets covering both morphological tissue-region and cellular/nuclear segmentation tasks.
Our method leverages attention maps of foundation models as pixel-wise features, which are then classified using a machine learning algorithm, XGBoost, enabling fast, interpretable, and model-agnostic evaluation without fine-tuning.
% We benchmark 10 foundation models on four datasets for
% histopathological image segmentation. 
% The approach utilizes the attention maps of foundation models to classify pixel-level features using the machine learning algorithm, XGBoost. 
% Moreover, our studies show that features extracted from different foundation models can be concatenated to deliver much superior performance. 
We show that the vision language foundation model, CONCH performed the best across datasets when compared to vision-only foundation models, with PathDino as close second. 
Further analysis shows that models trained on distinct histopathology cohorts capture complementary morphological representations, and concatenating their features yields superior segmentation performance.
% Our experiments reveal that foundation models trained on distinct cohorts learn complementary features to predict the same label, and their features can be concatenated to outperform individual models. 
Concatenating features from CONCH, PathDino and CellViT outperformed individual models across all the datasets by 7.95\% (averaged across the datasets), suggesting that ensembles of foundation models can better generalize to diverse histopathological segmentation tasks.
\end{abstract}

\section{Introduction}
\label{sec:intro}

Semantic segmentation is a fundamental step in computational pathology, enabling the precise delineation of glands, nuclei, and tissue regions that underpin diagnostic and prognostic workflows. Conventionally, training a model for semantic segmentation
involves creating a large amount of image labels, which present a significant challenge due to the labour intensive annotations. Recent histopathology specific variants, namely Virchow \cite{vorontsov2024virchowmillionslidedigitalpathology}, Phikon \cite{Filiot2023.07.21.23292757}, UNI \cite{Chen2024}  trained on large collections of whole slide images (WSIs) using self-supervised or multimodal learning objectives, have proven to be efficient feature extractors. The representations learned during the pretraining can be effectively transferred to the downstream task of semantic segmentation, using substantially fewer labeled data and computational resources. However, segmentation performance varies widely across foundation models, depending on their architecture, pretraining data, and learned representation highlighting the need for systematic benchmarking. Despite their potential, exploiting these foundation models (FMs) for pixel-wise segmentation remains underexplored. For instance, \cite{gildenblat2024segmentationfactorizationunsupervisedsemantic}  proposed an unsupervised segmentation approach via activation factorization while \cite{vitoria2025a} proposed a benchmarking methodology by appending a neural network decoder to utilize the output feature vector of FM's for supervised segmentation. A comprehensive review of such methods was presented in \cite{zhou2024imagesegmentationfoundationmodel}.

In this work, we address the lack of systematic benchmarking by proposing an approach to benchmark and utilize FMs for semantic segmentation. Our method leverages the attention maps of foundation models as pixel-level features, which are then classified using XGBoost \cite{Chen_2016}, enabling efficient segmentation without additional model fine-tuning.
% We propose a method which utilizes the attention maps of FMs to classify each pixel into a predefined class using XGBoost. 
We benchmark ten pathology-specific foundation models across multiple public datasets, quantifying their segmentation performance, analyzing their complementary strengths, and offering insights to guide model selection and future development in digital pathology. The key contributions of this work are summarized below.
% We present a comprehensive benchmarking study evaluating ten pathology-specific foundation models across multiple public datasets. This work aims to quantify their effectiveness for segmentation tasks, highlight their strengths and limitations, and provide practical insights to guide model selection and future development in digital pathology. 

% We benchmark numerous FMs at their effectiveness in addressing semantic segmentation tasks in histopathology. We experiment with several datasets covering cellular and morphological representations. 
% This work aims to quantify their effectiveness for segmentation tasks, highlight their strengths and limitations, and provide practical insights to guide model selection and future development in digital pathology.
\begin{figure*}[t]
\includegraphics[width=\textwidth]{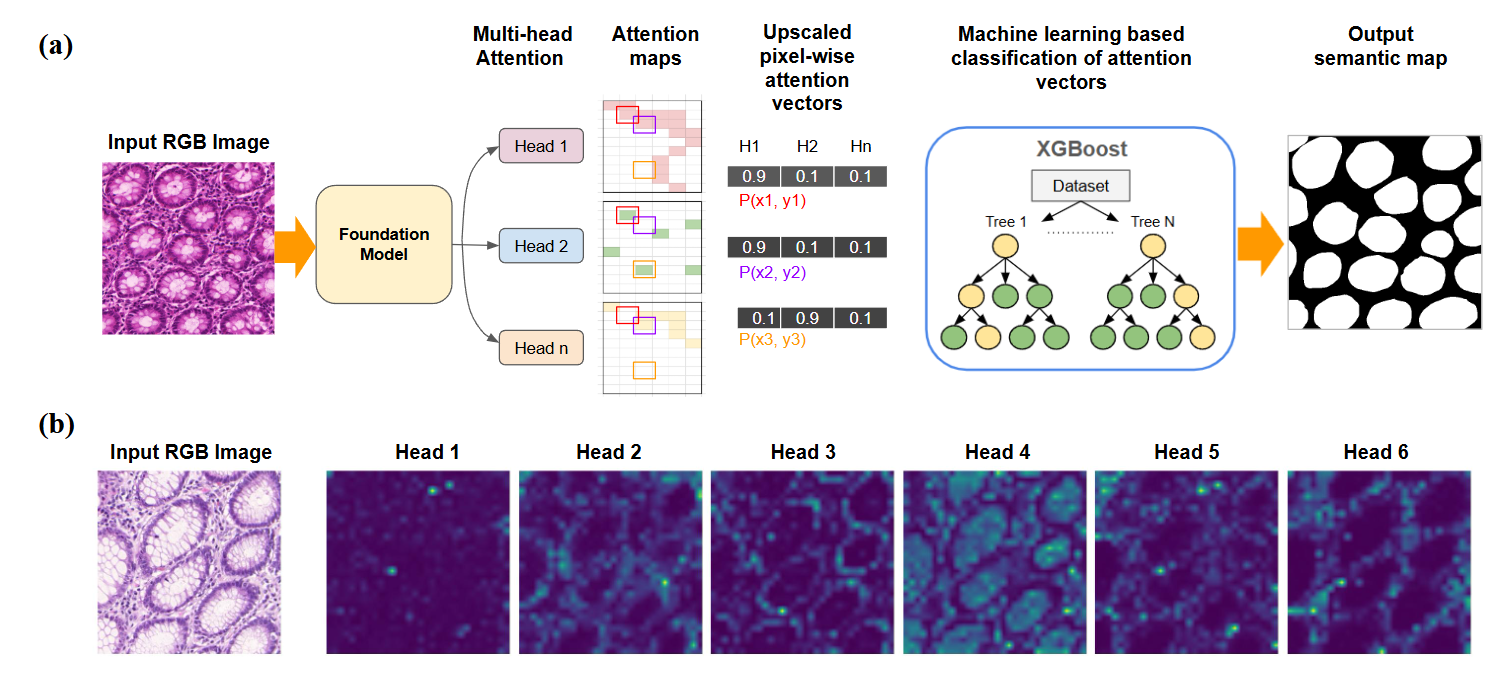}
\caption{(a)Overview of proposed method (b)Visualization of attention head outputs of PathDino on a sample patch from GlaS.} \label{fig1}
\end{figure*}

 \begin{table*}[]
 \small
\caption{Overview of FMs. Comparison by backbone, pretraining method, parameters and pretraining data source.}\label{tab1}
\resizebox{\textwidth}{!}{%
\begin{tabular}{cccccccc}
\hline
Foundation Models & Backbone           & Training method                     & WSIs in Pretraining dataset & Parameters  & Origin of WSI(s)                           \\ \hline
% Ctranspath        & Swin T/14              & Moco V3                             &                        &                     & 32.2          & 16 million  & TGCA, PAIP                                 \\ 
Virchow    \cite{vorontsov2024virchowmillionslidedigitalpathology}       & ViT-H                  & DINO v2                                       & 1.5 M   & 632 M & proprietary (from MSKCC)                   \\ 
Phikon  \cite{Filiot2023.07.21.23292757}          & ViT-B                  & iBOT                              & 6.1 k          & 86 M  &   TGCA                                         \\ 
UNI  \cite{Chen2024}             & ViT-L                  & DINO v2                                  & 100 k          &   303 M          & proprietry                                 \\ 
HIPT   \cite{chen2022scalingvisiontransformersgigapixel}           & ViT-S                  & DINO                                             & 11k           &    22 M         & TGCA                                       \\ 
Lunit-DINO   \cite{kang2022benchmarking}     & ViT-S                  & DINO                                              & 36 k         &      22 M       & TGCA \& proprietry                            \\ 
PathDino  \cite{alfasly2024rotationagnosticimagerepresentationlearning}        & ViT Custom             & DINO                                            & 11.8 k         & 9 M   & TGCA                                       \\ 
CellViT    \cite{hörst2023cellvitvisiontransformersprecise}       & ViT-S                  & Supervised                                   &            11 k   &      46 M        &     TGCA                                      \\ 
% UNI2              & ViT H/14               & DINO v2                             & 24                     & 16x16               & 350k          &             & TGCA/CPTAC/PANDA                           \\ 
Phikon-v2  \cite{filiot2024phikonv2largepublicfeature}       & ViT-L                  & DINO v2                                      & 58.4 k         & 307 M & PANCAN-XL (TCGA, CPTAC, GTEx, proprietary) \\ 

Virchow2 \cite{zimmermann2024virchow2scalingselfsupervisedmixed}         & ViT-H with 4 registers & DINO v2                                      & 3.1 M   & 632 M & proprietary (from MSKCC)                   \\ 
CONCH   \cite{lu2023visuallanguagefoundationmodelcomputational}          & ViT-B                  & iBOT \& vision-language pretraining        & 21 k          & 86 M  &              PMC-OA                              \\ 
% SAM               & ViT B                  & Supervised                          & 256                    & 64x64               &               &             &                                            \\ 
% MedSAM            & ViT B                  & Supervised                          & 256                    & 64x64               &               &             &                                            \\ 
\end{tabular}%
}
\end{table*}

\begin{enumerate}
\item We present a comprehensive benchmark study of a range of foundation models for dense pixel-wise segmentation tasks in the histopathology domain.
\item We demonstrate that concatenation of attentions from multiple foundation models improves segmentation performance, highlighting the property of FMs in learning complementary morphological representations.
\end{enumerate}

\makeatletter
\newcommand{\showfontsize}{The current font size is \f@size pt\par}
\makeatother

% \begin{document}
% Normal text: \showfontsize

% {\small Small text: \showfontsize}

% {\Large Large text: \showfontsize}
% \end{document}

% \small
% \usepackage{float}

% \makeatletter
% % \newcommand{\showfontsize}{The current font size is \f@size pt\par}
% \makeatother

% \begin{document}
% Normal text: \showfontsize

% {\small Small text: \showfontsize}

% {\Large Large text: \showfontsize}
% \end{document}

\begin{table*}[h]
\centering
\caption{Performance of foundation models across all datasets and tissue classes. Three best performing models are \textbf{boldfaced}, best performing model is also \underline{\textbf{underlined}}. Models are ordered according to their mean rank across the datasets.}\label{tab2}
\resizebox{\textwidth}{!}{%
\begin{tabular}{c|c|c|c|cccccc}
\hline
Foundation Model & GlaS & OCELOT Tissue & LyNSeC 2 & \multicolumn{6}{c}{BCSS}                                                                                                                                              \\ \hline
                 &      &               &          & \multicolumn{1}{c|}{Tumor} & \multicolumn{1}{c|}{Stroma} & \multicolumn{1}{c|}{Inflammatory} & \multicolumn{1}{c|}{Necrosis} & \multicolumn{1}{c|}{Others} & Mean Dice 
                 \\ \hline

CONCH            & \underline{\textbf{0.87}} & \textbf{0.67}          & \textbf{0.39}     & \multicolumn{1}{c|}{0.63}  & \multicolumn{1}{c|}{0.73}   & \multicolumn{1}{c|}{0.35}         & \multicolumn{1}{c|}{0.80}     & \multicolumn{1}{c|}{0.58}   & \underline{\textbf{0.62}}      \\ \hline
PathDino         & 0.83 & \underline{\textbf{0.69}}          & \textbf{0.56}     & \multicolumn{1}{c|}{0.42}  & \multicolumn{1}{c|}{0.70}   & \multicolumn{1}{c|}{0.40}         & \multicolumn{1}{c|}{0.73}     & \multicolumn{1}{c|}{0.08}   & 0.47      \\ \hline
CellViT         & 0.77 & \textbf{0.69}          & \underline{\textbf{0.82}}     & \multicolumn{1}{c|}{0.42}  & \multicolumn{1}{c|}{0.74}   & \multicolumn{1}{c|}{0.50}         & \multicolumn{1}{c|}{0.74}     & \multicolumn{1}{c|}{0.21}   & \textbf{0.52}      \\ \hline
Phikon           & 0.84 & 0.62          & 0.06     & \multicolumn{1}{c|}{0.59}  & \multicolumn{1}{c|}{0.69}   & \multicolumn{1}{c|}{0.26}         & \multicolumn{1}{c|}{0.76}     & \multicolumn{1}{c|}{0.40}   & \textbf{0.54}      \\ \hline
Phikon-v2        & \textbf{0.85} & 0.59          & 0.05     & \multicolumn{1}{c|}{0.55}  & \multicolumn{1}{c|}{0.69}   & \multicolumn{1}{c|}{0.24}         & \multicolumn{1}{c|}{0.75}     & \multicolumn{1}{c|}{0.36}   & 0.52      \\ \hline
Virchow          & \textbf{0.86} & 0.53          & 0.12     & \multicolumn{1}{c|}{0.35}  & \multicolumn{1}{c|}{0.66}   & \multicolumn{1}{c|}{0.14}         & \multicolumn{1}{c|}{0.69}     & \multicolumn{1}{c|}{0.43}   & 0.45      \\ \hline
Virchow2        & 0.82 & 0.62          & 0.08     & \multicolumn{1}{c|}{0.32}  & \multicolumn{1}{c|}{0.65}   & \multicolumn{1}{c|}{0.15}         & \multicolumn{1}{c|}{0.71}     & \multicolumn{1}{c|}{0.32}   & 0.43      \\ \hline
UNI              & 0.83 & 0.62          & 0.05     & \multicolumn{1}{c|}{0.17}  & \multicolumn{1}{c|}{0.60}   & \multicolumn{1}{c|}{0.42}         & \multicolumn{1}{c|}{0.57}     & \multicolumn{1}{c|}{0.06}   & 0.36      \\ \hline
Lunit DINO       & 0.81 & 0.54          & 0.01     & \multicolumn{1}{c|}{0.42}  & \multicolumn{1}{c|}{0.66}   & \multicolumn{1}{c|}{0.32}         & \multicolumn{1}{c|}{0.72}     & \multicolumn{1}{c|}{0.32}   & 0.49      \\ \hline
HIPT             & 0.71 & 0.53          & 0.13     & \multicolumn{1}{c|}{0.17}  & \multicolumn{1}{c|}{0.60}   & \multicolumn{1}{c|}{0.42}         & \multicolumn{1}{c|}{0.57}     & \multicolumn{1}{c|}{0.06}   & 0.36    
% UNI 2            & 0.73 & 0.64          & 0.04     & \multicolumn{1}{c|}{0.39}  & \multicolumn{1}{c|}{0.71}   & \multicolumn{1}{c|}{0.24}         & \multicolumn{1}{c|}{0.72}     & \multicolumn{1}{c|}{0.33}   & 0.48      \\ \hline
\end{tabular}%
}
\label{table:res}
\end{table*} 
\section{Methodology}
\label{sec:format}

The DINO \cite{caron2021emergingpropertiesselfsupervisedvision} framework demonstrated a surprising phenomenon that object segmentation can be obtained from the self-attention of class token (CLS) in the last attention layer of a transformer architecture. This property is resembled across different self supervised methods \cite{caron2021emergingpropertiesselfsupervisedvision}. DINO also showed that different attention heads attend to different semantic regions of an image. These properties are not shown by a fully supervised vision transformer (ViT) architecture both qualitatively and quantitatively \cite{caron2021emergingpropertiesselfsupervisedvision}. Self-supervised ViT features explicitly contain the scene layout and, in particular, object boundaries.

% \begin{figure}[htb]

% \begin{minipage}[b]{1.0\linewidth}
%   \centering
%   \centerline{\includegraphics[width=8.5cm]{example-image}}
% %  \vspace{2.0cm}
%   \centerline{(a) Result 1}\medskip
% \end{minipage}
% %
% \begin{minipage}[b]{.48\linewidth}
%   \centering
%   \centerline{\includegraphics[width=4.0cm]{example-image}}
% %  \vspace{1.5cm}
%   \centerline{(b) Results 3}\medskip
% \end{minipage}
% \hfill
% \begin{minipage}[b]{0.48\linewidth}
%   \centering
%   \centerline{\includegraphics[width=4.0cm]{example-image}}
% %  \vspace{1.5cm}
%   \centerline{(c) Result 4}\medskip
% \end{minipage}
% %
% \caption{Example of placing a figure with experimental results.}
% \label{fig:res}
% %
% \end{figure}

Our proposed method leverages the self-attention of class token (CLS) in the last attention layer of a transformer architecture as shown in Fig.~\ref{fig1}. 

% The self-attention mechanism in each transformer block learns the relationship between distinct tokens of an image using multiple attention heads. 

The self-attention mechanism in transformer-based foundation models captures contextual relationships between tokens through multiple attention heads, encoding both local and global spatial dependencies. In our proposed method, we upsample the token level attention maps from the final transformer block to the input image dimensions using bilinear interpolation, resulting in a pixel-level feature map. These pixel-level features are then used to train a supervised classifier, XGBoost \cite{Chen_2016}. The trained classifier predicts class probabilities for each pixel, yielding the final semantic segmentation map.
% We train a supervised machine learning (ML) algorithm XGBoost \cite{Chen_2016} on the extracted pixel-level features. Output semantic map is obtained by classifying these pixel level features into predefined classes.

% Finally, we apply morphological operations on the predictions of XGBoost to smoothen the boundaries and remove small pixelated outputs. 
In contrast to recent decoder based approaches \cite{vitoria2025a}, we propose a decoder-free benchmarking framework leveraging XGBoost. Decoder fine-tuning can introduce training bias due to different encoder architectures, data-size sensitivity, and optimization variance, obscuring the intrinsic quality of learned representations. In contrast, XGBoost operates directly on frozen features offering a lightweight, non-parametric alternative that enables efficient and interpretable cross-model comparisons.

Studies have shown that FMs trained on distinct cohorts learn complementary representations \cite{neidlinger2024benchmarkingfoundationmodelsfeature}. Hence, we experimented with concatenating attention maps from multiple foundation models. 
% Concatenation of the attention maps preserves the features extracted from each individual model unlike ensembling which results in an averaging effect. 

% Studies have shown ref(https://arxiv.org/abs/2012.15840) that direct use of transformer encoder features for pixel-level prediction performs poorly and a sophisticated decoder is required to recover spatial information. Hence, we selected XGBoost due to its superior performance across datasets compared to other machine learning methods and linear probing. 

% Previous studies \cite{zheng2021rethinkingsemanticsegmentationsequencetosequence} have shown that transformer encoder features alone are insufficient for pixel-level predictions without an effective decoder. To address this, we employed XGBoost, which consistently outperforms other machine learning approaches and linear probing across datasets.

\section{Experiment Details}
\label{sec:pagestyle}

In this study, we benchmarked ten pathology specific foundation models (Table \ref{tab1}) pre-trained on a large collection of images. We employed 4 public datasets (Table \ref{tab:datasets}) to evaluate their performance on semantic segmentation tasks namely,  Gland Segmentation in Colon Histology Images (GlaS) \cite{sirinukunwattana2016glandsegmentationcolonhistology}, Overlapped Cell on Tissue Dataset for Histopathology (OCELOT) Tissue \cite{ryu2023ocelotoverlappedcelltissue}, LyNSeC (lymphoma nuclear segmentation and classification) 2 \cite{naji2024holynet} and Breast Cancer Semantic Segmentation (BCSS) dataset \cite{10.1093/bioinformatics/btz083}. Datasets were split according to the train–test partitions provided by the dataset authors whenever available; otherwise, random splitting was applied (Table \ref{tab:datasets}).

Each model was evaluated using its native pretraining input resolution; input patches were resized accordingly prior to feature extraction to ensure consistency with model-specific training scales.

We implemented our method using Python (version 3.12), Pytorch (version 2.5) and a workstation with 1 NVIDIA RTX A6000 GPU with 48 GB memory. We trained XGBoost
 (version 2.1.1) with the “hist” tree method for 100 boosting rounds. Model performance was evaluated using the Dice score as the primary metric.
 
% \begin{table*}[]
% \begin{tabular}{llllll}
% Dataset       & Magnification & Patch shape & No. of Patches in Train set & No. of Patches in Test Set & Dataset Split   \\
% Glas          & 20x           & 776x524     & 85                          & 82                         & Given by Author \\
% OCELOT Tissue & 10x           & 1024x1024   & 320                         & 80                         & Random          \\
% Lynsec 2      & 40x           & 512x512     & 224                         & 56                         & Random          \\
% BCSS          & 20x ,40x      & 1024x1024   & 1908                        & 977                        & Given by Author
% \end{tabular}
% \end{table*}

% \begin{table*}[]
% \begin{tabular}{l|l|l|l|l|l}
% Dataset       & Magnification & Patch shape & No. of Patches in Train set & No. of Patches in Test Set & Dataset Split   \\\hline
% Glas          & 20x           & 776x524     & 85                          & 82                         & Given by Author \\
% OCELOT Tissue & 10x           & 1024x1024   & 320                         & 80                         & Random          \\
% Lynsec 2      & 40x           & 512x512     & 224                         & 56                         & Random          \\
% BCSS          & 20x ,40x      & 1024x1024   & 1908                        & 977                        & Given by Author
% \end{tabular}
% \end{table*}

\begin{table}[h]
\centering
\caption{Summary of the datasets used for benchmarking.}
\resizebox{\columnwidth}{!}{%
\begin{tabular}{l|l|l|l|l|l}
\hline
Dataset & Magnification & Patch shape & \# Train Patches & \# Test Patches & Split \\
\hline
GlaS & 20x & 776×524 & 85 & 82 & Given by author \\
OCELOT Tissue & 10x & 1024×1024 & 320 & 80 & Random \\
LyNSeC 2 & 40x & 512×512 & 224 & 56 & Random \\
BCSS & 20x, 40x & 1024×1024 & 1908 & 977 & Given by author \\
\hline
% \small
\end{tabular}%
}
\label{tab:datasets}
\end{table}

\begin{figure}[h]

\begin{minipage}[b]{1.0\linewidth}
  \centering
  \centerline{\includegraphics[width=8.5cm]{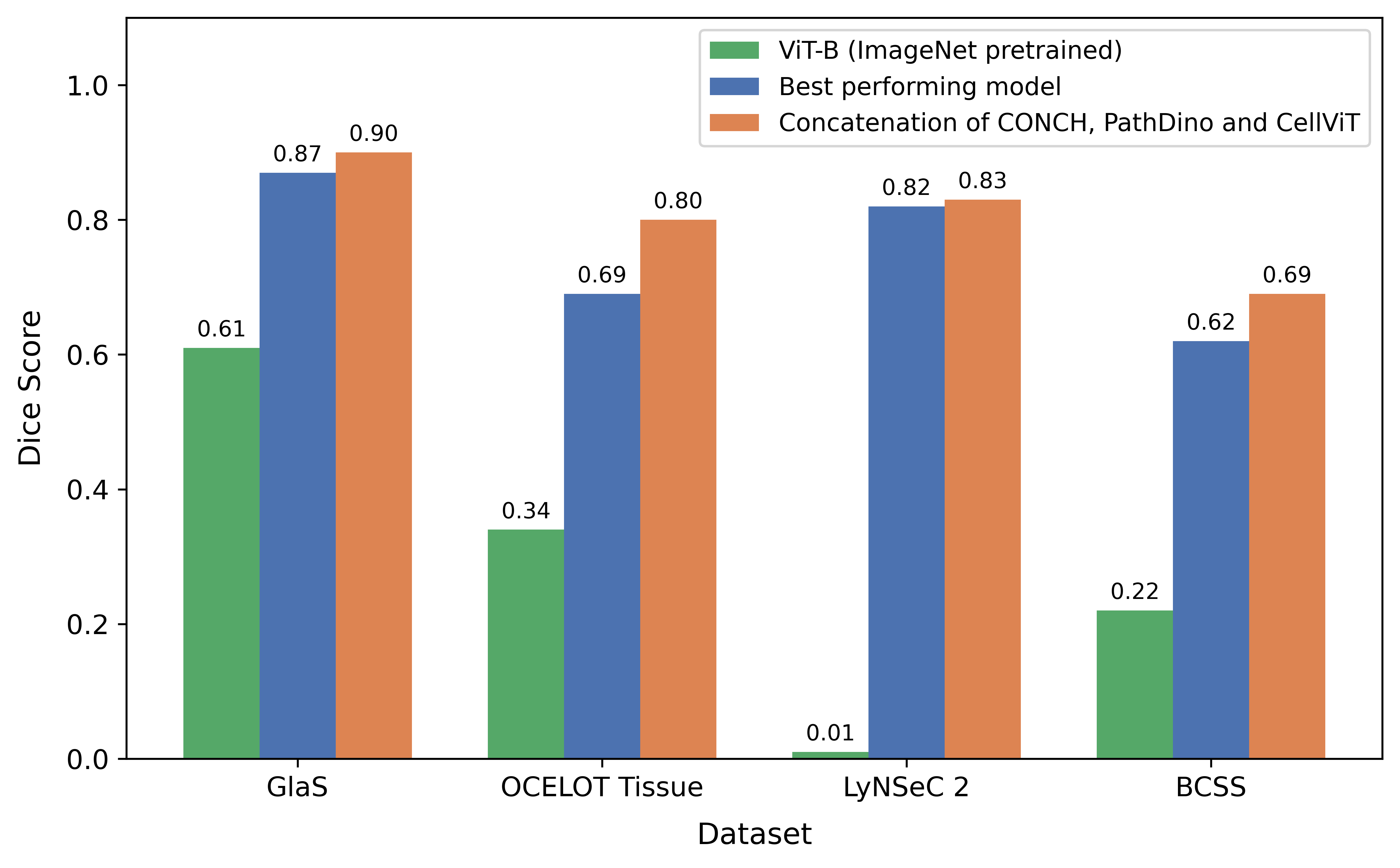}}
%  \vspace{2.0cm}
  % \centerline{(a) Result 1}\medskip
\end{minipage}
\caption{Comparison between  ViT-B, the best individual FM and the concatenated feature setup (CONCH + PathDino + CellViT) on each dataset. For the BCSS dataset, the Dice score averaged across the five tissue classes is reported.}
\label{fig:res}
\end{figure}
 
\section{Results and Discussion}
\label{sec:typestyle}

Table \ref{table:res} shows the results of benchmarking of foundation models across the four datasets. Among all the models, CONCH performs the best across the datasets (Figure \ref{fig:res2}), demonstrating its superior capability in representing both morphological and contextual information likely due to its vision-language pretraining paradigm. This reinforces the hypothesis that multimodal pretraining can effectively enhance feature extraction.

% Furthermore, as shown in Figure~\ref{fig:res2}, all histopathology-specific foundation models consistently outperform the ViT-B model pretrained on ImageNet, underscoring the importance of domain-specific pretraining for dense prediction tasks in pathology.

Following CONCH, PathDino and CellViT  showed strong generalization across the datasets. PathDino, despite utilizing only five transformer blocks, gave consistent performance in comparison to other FMs with deeper architectures (e.g., Virchow, Phikon, and UNI), highlighting the robustness of its Rotation-agnostic training strategy.  Notably, Virchow2, despite being trained on over three million WSIs, showed moderate performance, indicating that scale alone may not guarantee effective downstream segmentation. Instead, the diversity and granularity of pretraining data appear to play a critical role. Similar to \cite{vitoria2025a}'s observation, our results indicate that advanced versions such as Phikon-v2 and Virchow2 do not consistently outperform their predecessors, implying that improvements in classification performance do not necessarily translate to gains in pixel-level segmentation tasks.
% This indicates that simply having deeper architectures and more training data may not be the best recipe for improving the performance of VFMs.

% Unlike the other three datasets, the LYnsec datasets  tiles differ from the H&E appearance most encoders were pretrained on. The task is therefore a good cross-stain generalisation test.

% Second-generation checkpoints with larger
% architectures or extended training datasets (e.g. UNI2-H vs. UNI, Virchow-2 vs. Virchow-1)
% underperform their predecessors, possibly because extensive classification-oriented fine-tuning weakens the positional correlations in embeddings, crucial for pixel-level tasks.

% Histopathology focused foundation models, including UNI, Phikon, and Virchow, along with their variants, exhibited relatively poor performance on the LyNSeC 2 dataset, revealing potential limitations in capturing fine-grained cellular-level features.

% Histopathology focused foundation models, including 
UNI, Phikon, and Virchow, along with their variants, exhibited relatively poor performance on the LyNSeC 2 dataset, revealing potential limitations in capturing finegrained cellular representations. In contrast, CellViT achieved the best performance as can be seen in Figure \ref{fig:res2}, since it's architecture is optimized for cellular segmentation tasks. In this study, we utilized the upsampled final embedding (of spatial size 256×256 with 64 channels) of CellViT  as the pixel-level feature map.

While some of our overall observations are consistent with those reported by \cite{vitoria2025a}, certain discrepancies in the relative ranking of foundation models are evident. These differences may arise from variations in dataset scale and evaluation design, the prior study utilized only 15 patches in training and 10 for testing, whereas our work leverages the complete public datasets for benchmarking.

% The smaller sample size in the earlier work may have amplified data-specific biases and limited the statistical robustness of model comparisons.

Interestingly, concatenation of attention maps from multiple FMs yields significant performance improvements over individual models (Figure \ref{fig:res}). All histopathology-specific foundation models consistently outperform the ViT-B model pretrained on ImageNet (Figure \ref{fig:res}), further emphasizing the importance of domain-specific pretraining for histopathological tasks. The concatenation of attention maps of CONCH, PathDino, and CellViT  improved the performance across all the datasets, underscoring the complementary nature of representations learned by models trained independently. Unlike ensemble averaging, concatenation preserves distinct feature manifolds learned during pretraining, allowing downstream classifier XGBoost to exploit orthogonal discriminative cues.

% \begin{figure}[h]
% \begin{minipage}[b]{1.0\linewidth}
%   \centering
%   \centerline{\includegraphics[width=8.5cm]{image.png}}
% %  \vspace{2.0cm}
%   % \centerline{(a) Result 1}\medskip
% \end{minipage}
% \caption{Segmentation outputs of CONCH, PathDino and CellViT on different datasets. Post processed to smoothen the boundaries and remove pixelated outputs.}
% \label{fig:res2}
% \end{figure}

\section{Conclusion}
\label{sec:typestyle}

 In this work, we introduced a decoder-free benchmarking approach and comprehensively benchmarked ten foundation models for histopathological image segmentation. Our results concluded that CONCH, PathDino and CellViT give the best performance across the datasets, whereas newer and larger models such as Phikon-v2 and Virchow2 despite being trained on millions of WSIs do not consistently outperform their predecessors. This indicates that classification performance and pretraining scale do not necessarily translate to superior segmentation ability in downstream tasks. Furthermore, we established different pretraining objectives lead to learning complementary representations, underscoring the potential of ensemble or hybrid strategies. We anticipate that our work will serve as a springboard, sparking future exploration in developing and leveraging foundation models for semantic segmentation tasks.

\section{COMPLIANCE WITH ETHICAL STANDARDS}
\label{sec:compliance}

This research study was conducted retrospectively using human subject data made available in open access. Ethical approval was not required as confirmed by the license attached with the open access data.

\ninept

\makeatletter
\renewenvironment{thebibliography}[1]
     {\section{References}%
      \list{\@biblabel{\arabic{enumi}}}%
           {\settowidth\labelwidth{\@biblabel{#1}}%
            \leftmargin\labelwidth
            \advance\leftmargin\labelsep
            \usecounter{enumi}%
            \setlength{\itemsep}{0pt} % <-- no extra vertical space
            \setlength{\parsep}{0pt}%
            \setlength{\parskip}{0pt}}%
      \def\newblock{\hskip .11em plus .33em minus .07em}%
      \sloppy\clubpenalty4000\widowpenalty4000%
      \sfcode`\.=1000\relax}
     {\endlist}
\makeatother

\bibliographystyle{IEEEbib}
\bibliography{strings,refs_full}

\end{document}